\documentclass{article}

    \PassOptionsToPackage{numbers, compress}{natbib}

\usepackage[preprint]{neurips_2023}

\usepackage[utf8]{inputenc} 
\usepackage[T1]{fontenc}    
\usepackage{hyperref}       
\usepackage{url}            
\usepackage{booktabs}       
\usepackage{amsfonts}       
\usepackage{nicefrac}       
\usepackage{microtype}      
\usepackage{xcolor}         
\usepackage{wrapfig}
\usepackage{graphicx}    
\usepackage{multirow}
\usepackage{tabularx}
\usepackage{colortbl}
\usepackage{amsmath}
\definecolor{Gray}{gray}{0.9}
\definecolor{Gray2}{gray}{0.95}
\usepackage{pifont}
 \newcommand{\cmark}{\ding{51}}%
\newcommand{\xmark}{\ding{55}}%

\title{MiniGPT-v2: Large Language Model As a Unified Interface for Vision-Language Multi-task Learning}

\author{Jun Chen\textsuperscript{\rm 1,\rm 2\thanks{Work partially done during the internship at Meta AI}} \space Deyao Zhu\textsuperscript{\rm 1} \space   Xiaoqian Shen\textsuperscript{\rm 1} \space Xiang Li\textsuperscript{\rm 1} \space  Zechun Liu\textsuperscript{\rm 2} \space Pengchuan Zhang\textsuperscript{\rm 2}  \\ \textbf{Raghuraman Krishnamoorthi}\textsuperscript{\rm 2} \space \textbf{Vikas Chandra}\textsuperscript{\rm 2} \space  \textbf{Yunyang Xiong}\textsuperscript{\rm 2\thanks{Equal last author}} \space   \textbf{Mohamed Elhoseiny}\textsuperscript{\rm 1\footnote[2]{ Equal last author }} \\
\textsuperscript{\rm 1}{King Abdullah University of Science and Technology (KAUST)} \\
\textsuperscript{\rm 2}{Meta AI Research}
}

\begin{document}

\maketitle

\begin{abstract}  

Large language models have shown their remarkable capabilities as a general interface for various language-related applications. Motivated by this, we target to build a unified interface for completing many vision-language tasks including image description, visual question answering, and visual grounding, among others. The challenge is to use a single model for performing diverse vision-language tasks effectively with simple multi-modal instructions. 
Towards this objective, we introduce MiniGPT-v2, a model that can be treated as a unified interface for better handling various vision-language tasks. We propose using unique identifiers for different tasks when training the model. These identifiers enable our model to better distinguish each task instruction effortlessly and also improve the model learning efficiency for each task. After the three-stage training, the experimental results show that MiniGPT-v2 achieves strong performance on many visual question-answering and visual grounding benchmarks
compared to other vision-language generalist models. Our model and codes are available at \url{https://minigpt-v2.github.io/}.


\end{abstract}

\section{Introduction}

Multi-modal Large Language Models (LLMs) have emerged as an exciting research topic with a rich set of applications in vision-language community, such as visual AI  assistant, image captioning, visual question answering (VQA), and referring expression comprehension (REC). A key feature of multimodal large language models is that they can inherit advanced capabilities (e.g., logical reasoning, common sense, and strong language expression) from the LLMs~\citep{chatgpt,llama,llama2,vicuna2023}. When tuned with proper vision-language instructions, multi-modal LLMs, specifically vision-language models, demonstrate strong capabilities such as producing detailed image descriptions, generating code, localizing the visual objects in the image, and even performing multi-modal reasoning to better answer complicated visual questions~\citep{zhu2023minigpt,llava,mplugowl,wang2023visionllm,shikra,instructblip, chatcaptioner,chen2023video,mindstorms}. This evolution of LLMs enables interactions of visual and language inputs across communication with individuals and has been shown quite effective for building visual chatbots.

However, learning to perform  multiple vision-language tasks effectively and formulating their corresponding multi-modal instructions present considerable challenges due to the complexities inherent among different tasks. For instance, given a user input \textit{``tell me the location of a person"}, there are many ways to interpret and respond based on the specific task. In the context of the referring expression comprehension task, it can be answered with one bounding box location of the person. For the visual question-answering task, the model might describe their spatial location using human natural language.
For the person detection task, the model might identify every spatial location of each human in a given image. To alleviate this issue and towards a unified approach, we propose a task-oriented instruction training scheme to reduce the multi-modal instructional ambiguity, and a vision-language model, MiniGPT-v2. Specifically, we provide a unique task identifier token for each task. For example, we provide a \textit{[vqa]} identifier token for training all the data samples from the visual question answering tasks. In total, we provide six different task identifiers during the model training stages.


Our model, MiniGPT-v2, has a simple architecture design. It directly takes the visual tokens from a ViT vision encoder~\citep{fang2022eva} and project them into the feature space of a large language model~\citep{llama2}. For better visual perception, we utilize higher-resolution images (448x448) during training. But this will result in a larger number of visual tokens. To make the model training more efficient, we concatenate every four neighboring visual tokens into a single token, reducing the total number by 75\%. Additionally, we utilize a three-stage training strategy to effectively train our model with a mixture of weakly-labeled, fine-grained image-text datasets, and multi-modal instructional datasets, with different training focus at each stage.

\begin{figure}[t]
    \centering
    \includegraphics[width=0.9\textwidth]{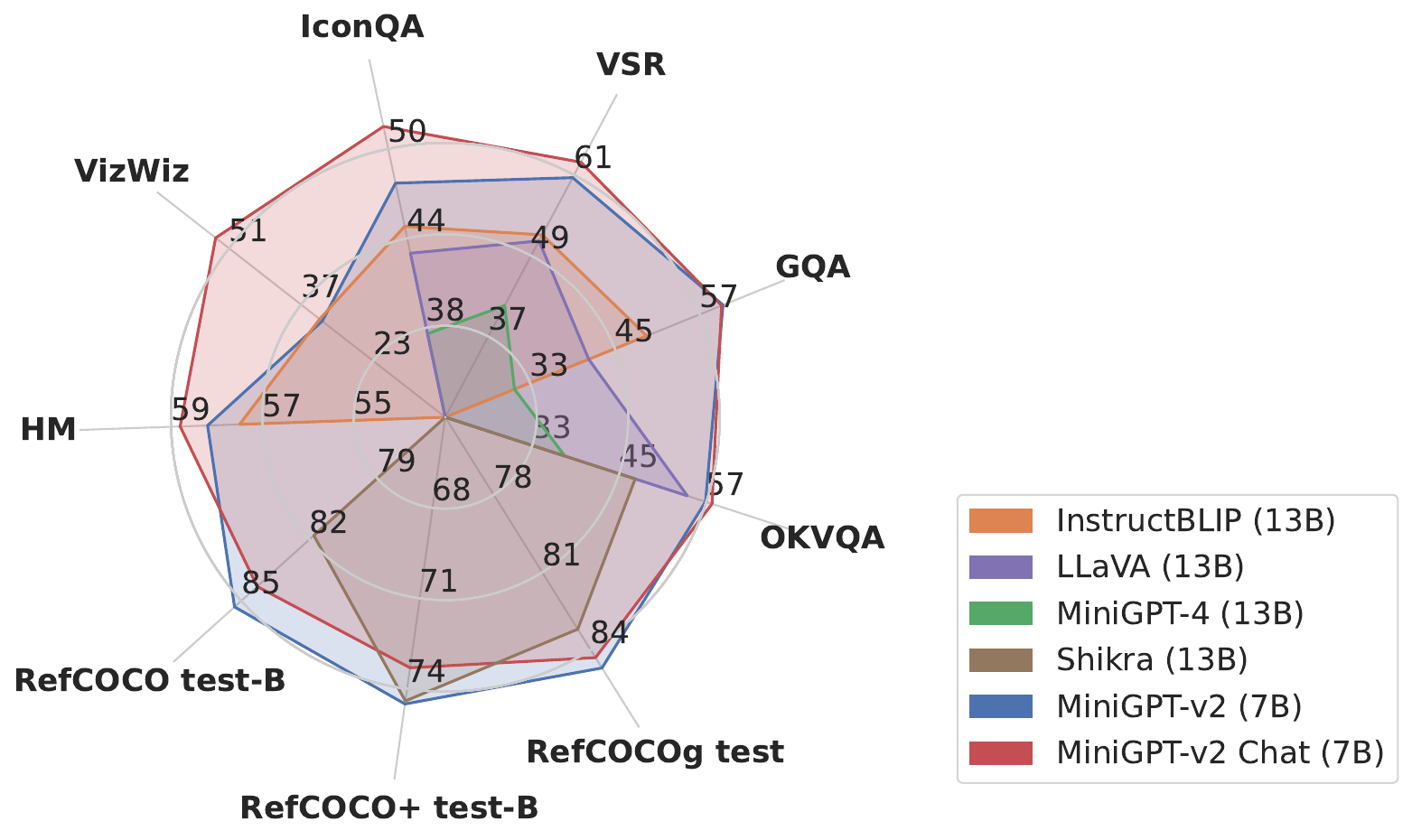}
    \caption{Our MiniGPT-v2 achieves state-of-the-art performances on a broad range of vision-language tasks compared with other generalist models.}
    \label{fig:radar1}
\end{figure}

To evaluate the performance of our model, we conducted extensive experiments on diverse vision-language tasks, including (detailed) image/grounded captioning, vision question answering, and visual grounding. The results demonstrate that our MiniGPT-v2 can achieve SOTA or comparable performance on diverse benchmarks compared to previous vision-language generalist models, such as MiniGPT-4~\citep{zhu2023minigpt}, InstructBLIP~\citep{instructblip}, LLaVA~\citep{llava} and Shikra~\citep{shikra}. For example, our MiniGPT-v2 outperforms MiniGPT-4 by 21.3\%, InstructBLIP by 11.3\%, and LLaVA by 11.7\% on the VSR benchmark~\citep{liu2023visual}, and it also performs better than the previously established strong baseline, Shikra, in most validations on RefCOCO, RefCOCO+, and RefCOCOg. Our model establishes new state-of-the-art results on these benchmarks among vision-language generalist models, shown in Fig. \ref{fig:radar1}. 

\section{Related Work}

We briefly review relevant works on advanced large language models and multi-modal LLMs for visual aligning.

\textbf{Advanced Large Language Models (LLMs).} 
Early-stage models such as GPT-2~\citep{gpt2} and BERT~\citep{bert} are foundation models trained on web-scale text datasets, marking a breakthrough in the NLP field. Following the success of foundation models, LLMs with higher capacity and increased training data are developed, including GPT-3~\citep{gpt3}, Megatron-turing NLG~\citep{smith2022using}, PaLM~\citep{chowdhery2022palm}, Gopher~\citep{rae2021scaling}, Chinchilla~\citep{hoffmann2022training}, OPT~\citep{zhang2022opt}, and BLOOM~\citep{scao2022bloom}. Most recently, the efforts have been focused on refining LLMs to work effectively with human instruction and feedback. Representative works in this direction are InstructGPT~\citep{instructGPT} and ChatGPT~\citep{chatgpt}, which demonstrate strong capabilities such as answering a diverse range of language questions, engaging in conversations with humans, and learning to perform complex tasks like writing refinement and coding assistant.

Concurrent with these advancements of LLMs is the rise of LLaMA~\citep{llama} language models. To enable human instruction following abilities similar to ChatGPT, some works attempt to finetune the LLaMA model with additional high-quality instruction datasets~\citep{sharegpt}. Examples of these models include Alpaca~\citep{alpaca}, Vicuna ~\citep{vicuna2023}, and MPT~\citep{mpt}. Some other open-sourced language models that learned from the human feedback data, such as Falcon~\citep{falcon} and LLaMA-2~\citep{llama2}, have also been introduced to the NLP community with impressive performance.


\textbf{Visual Aligning with LLMs.} With the remarkable generalization abilities of LLMs, interesting studies have extended LLMs to multi-modal domains by aligning visual inputs with LLMs. Early works such as VisualGPT~\citep{visualgpt} and Frozen~\citep{tsimpoukelli2021multimodal} used pre-trained language models to improve vision-language models on image captioning and visual question answering. This initial exploration paved the way for subsequent vision-language research such as Flamingo~\citep{alayrac2022flamingo} and BLIP-2~\citep{blip2}. More recently, GPT-4 has been released and demonstrates many advanced multi-modal abilities, e.g.,  generating website code based on handwritten text instructions. Those demonstrated capabilities inspired other vision-language LLMs, including MiniGPT-4~\citep{zhu2023minigpt} and LLaVA~\citep{llava}, which align the image inputs with a large language model, Vicuna~\cite{vicuna2023}, using proper instructional tuning. These vision-language models also showcase many advanced multi-modal capabilities after the alignment. Recent works, such as Vision-LLM~\citep{wang2023visionllm}, Kosmos-2~\citep{peng2023kosmos}, Shikra~\citep{shikra}, and our concurrent work, Qwen-VL~\citep{qwen}, also demonstrate that multi-model LLMs models can also perform visual grounding by generating the text format of bounding boxes through language model. 


\section{Method}

\begin{wrapfigure}{r}{0.5\textwidth}
\vspace{-1cm}
    \centering
    \includegraphics[width=0.5\textwidth]{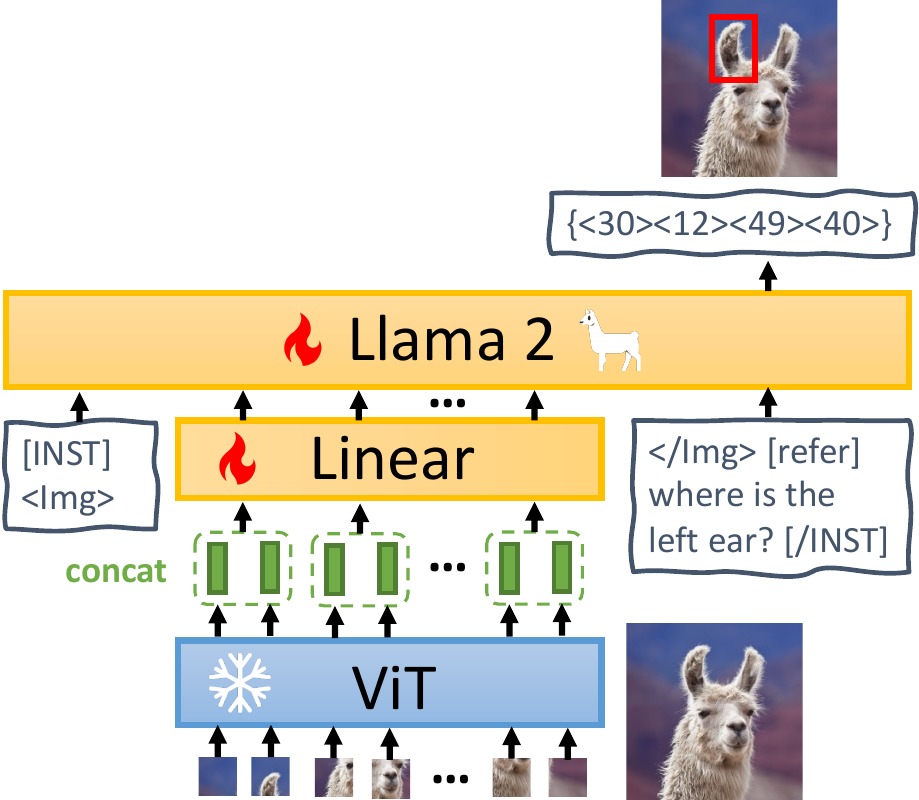}
    \caption{\textbf{Architecture of MiniGPT-v2.} The model takes a ViT visual backbone, which remains frozen during all training phases. We concatenate four adjacent visual output tokens from ViT backbone and project them into LLaMA-2 language model space via a linear projection layer.}
    \label{fig:arch}
\vspace{-2cm}
\end{wrapfigure}

We start by introducing our vision-language model, MiniGPT-v2, then discuss the basic idea of a multi-task instruction template with task identifiers for training, and finally adapt our task identifier idea to achieve task-oriented instruction tuning. 
\subsection{Model Architecture}


Our proposed model architecture, MiniGPT-v2, is shown in Fig. \ref{fig:arch}. It consists of three components: a visual backbone, a linear projection layer, and a large language model. We describe each component as follows:

\textbf{Visual backbone.} MiniGPT-v2 adapts the EVA~\citep{fang2022eva} as our visual backbone model backbone.  We freeze the visual backbone during the entire model training. We train our model with the image resolution 448x448, and we interpolate the positional encoding to scale with a higher image resolution.

\textbf{Linear projection layer.} We aim to project all the visual tokens from the frozen vision backbone into the language model space. However, for higher-resolution images such as 448x448, projecting all the image tokens results in a very long-sequence input (e.g., 1024 tokens) and significantly lowers the training and inference efficiency. Hence, we simply concatenate 4 adjacent visual tokens in the embedding space and project them together into one single embedding in the same feature space of the large language model, thus reducing the number of visual input tokens by 4 times. With this operation, our MiniGPT-v2 can process high-resolution images much more efficiently during the training and inference stage.

\textbf{Large language model.} MiniGPT-v2 adopts the open-sourced LLaMA2-chat (7B)~\citep{llama2} as the language model backbone. In our work, the language model is treated as a unified interface for various vision-language inputs. We directly rely on the LLaMA-2 language tokens to perform various vision-language tasks. For the visual grounding tasks that necessitate the generation of spatial locations, we directly ask the language model to produce textual representations of bounding boxes to denote their spatial positions.


\subsection{Multi-task Instruction Template} 
When training a single unified model for multiple different tasks such as visual question answering, image caption, referring expression, grounded image caption, and region identification, the multi-modal model might fail to distinguish each task by just aligning visual tokens to language models. For instance, when you ask ``Tell me the spatial location of the person wearing a red jacket?'', the model can either respond you the location in a  bounding box format (e.g., $<\text{X}_{left}><\text{Y}_{top}><\text{X}_{right}><\text{Y}_{bottom}>$) or describe the object location using natural language (e.g., upper right corner). To reduce such ambiguity and make each task easily distinguishable, we introduce task-specific tokens in our designed multi-task instruction template for training. We now describe our multi-task instruction template in more details. 

\textbf{General input format.} We follow the LLaMA-2 conversation template design and adapt it for the multi-modal instructional template. The template is denoted as follows,
\begin{center}
\textit{[INST] $<$Img$>$ $<$ ImageFeature$>$ $<$/Img$>$ [Task Identifier] Instruction [/INST]} 
\end{center}
In this template, \textit{[INST]} is considered as the user role, and \textit{[/INST]} is considered as the assistant role. We structure the user input into three parts. The first part is the image features, the second part is the task identifier token, and the third part is the instruction input. 

\textbf{Task identifier tokens.} Our model takes a distinct identifier for each task to reduce the ambiguity across various tasks. As illustrated in Table \ref{identifier}, we have proposed six different task identifiers for visual question answering, image caption, grounded image captioning, referring expression comprehension,  referring expression generation, and phrase parsing and grounding respectively. For vision-irrelevant instructions, our model does not use any task identifier token.


\begin{table*}[h!]
\centering 
\resizebox{\textwidth}{!}{
\begin{tabular}{@{}lc ccccc @{}} %
Tasks & VQA &   Caption & Grounded Caption & REC  & REG & Object Parsing and Grounding \\  
\toprule
Identifiers & [vqa] & [caption] &[grounding] &[refer]   & [identify] & [detection] \\
\end{tabular}
}
\caption{Task identifier tokens for 6 different tasks, including visual question answering, image captioning, grounded image captioning, referring expression comprehension (REC), referring expression generation (REG), and object parsing and grounding (where the model extracts objects from the input text and determines their bounding box locations).}
\label{identifier}
\end{table*}

\textbf{Spatial location representation.} For tasks such as referring expression comprehension (REC), referring expression generation (REG), and grounded image captioning, our model is required to identify the spatial location of the referred objects accurately. We represent the spatial location through the textual formatting of bounding boxes in our setting, specifically: ``$\{<\text{X}_{left}><\text{Y}_{top}><\text{X}_{right}><\text{Y}_{bottom}>\}$". Coordinates for X and Y are represented by integer values normalized in the range [0,100]. $<\text{X}_{left}>$ and $<\text{Y}_{top}>$ denote the x and y coordinate top-left corner of the generated bounding box, and $<\text{X}_{right}>$ and $<\text{Y}_{bottom}>$ denote the x and y coordinates of the bottom-right corner.


\subsection{Multi-task Instruction Training}
We now adapt our designed multi-task instruction template for instruction training. The basic idea is to take instruction with task-specific identifier token as input for task-oriented instruction training of MiniGPT-v2. When input instructions have task identifier tokens, our model will become more prone to multiple-task understanding during training. We train our model with task identifier instructions for better visual aligment in three stages. The first stage is to help MiniGPT-v2 build broad vision-language knowledge through many weakly-labeled image-text datasets, and high-quality fine-grained vision-language annotation datasets as well (where we will assign a high data sampling ratio for weakly-labeled image-text datasets). The second stage is to improve the model with only fine-grained data for multiple tasks. The third stage is to finetune our model with more multi-modal instruction and language datasets for answering diverse multi-modal instructions better and behaving as a multi-modal chatbot. The datasets used for training at each stage are listed in Table \ref{datasets}. 

\begin{table*}[h!]
\small
\centering 
\resizebox{\textwidth}{!}{
\begin{tabular}{@{}ll c c c @{}} %
\toprule
 Data types &   Dataset  &  Stage 1& Stage 2 &  Stage 3 \\ 
 \midrule
Weakly-labeled & GRIT-20M (REC and REG), LAION, CC3M, SBU & \cmark  & \xmark & \xmark \\
Grounded caption & GRIT-20M & \cmark  & \xmark & \xmark  \\
Caption &  COCO caption,  Text Captions  & \cmark  & \cmark & \cmark  \\
REC &  RefCOCO, RefCOCO+, RefCOCOg, Visual Genome & \cmark  & \cmark & \cmark   \\
REG & RefCOCO, RefCOCO+, RefCOCOg & \cmark  & \cmark & \cmark \\
VQA &  GQA, VQAv2, OCR-VQA, OK-VQA, AOK-VQA  & \cmark  & \cmark & \cmark\\
Multimodal instruction & LLaVA dataset, Flickr30k, Multi-task conversation & \xmark  & \xmark & \cmark\\
Langauge dataset & Unnatural Instructions & \xmark  & \xmark & \cmark\\
\bottomrule
\end{tabular}
}
\caption{The training datasets used for our model three-stage training.
}
\label{datasets}
\end{table*}

\textbf{Stage 1: Pretraining.} To have broad vision-language knowledge, our model is trained on a mix of weakly-labeled and fine-grained datasets. We give a high sampling ratio for weakly-labeled datasets to gain more diverse knowledge in the first-stage.

For the weakly-labeled datasets, we use LAION~\citep{laion}, CC3M~\citep{sharma2018conceptual}, SBU~\citep{sbu}, and GRIT-20M from Kosmos v2~\citep{peng2023kosmos} that built the dataset for referring expression comprehension (REC), referring expression generation (REG), and grounded image captioning. 

For fine-grained datasets, we use datasets like COCO caption~\citep{lin2014microsoft} and Text Captions~\citep{sidorov2019textcaps} for image captioning, RefCOCO~\citep{kazemzadeh2014referitgame}, RefCOCO+~\citep{yu2016modeling}, and RefCOCOg~\citep{mao2016generation} for REC. For REG, we restructured the data from ReferCOCO and its variants, reversing the order from phrase $\rightarrow$ bounding boxes to bounding boxes $\rightarrow$ phrase. For VQA datasets, our training takes a variety of datasets, such as GQA~\citep{hudson2019gqa}, VQA-v2~\citep{goyal2017making}, OCR-VQA~\citep{ocrvqa}, OK-VQA~\citep{marino2019ok}, and AOK-VQA~\citep{schwenk2022okvqa}.



\textbf{Stage 2: Multi-task training.} To improve the performance of MiniGPT-v2 on each task, we only focus on using fine-grained datasets to train our model at this stage. We exclude the weakly-supervised datasets such as GRIT-20M and LAION from stage-1 and update the data sampling ratio according to the frequency of each task. This strategy enables our model to prioritize high-quality aligned image-text data for superior performance across various tasks.

\textbf{Stage 3: Multi-modal instruction tuning.} Subsequently, we focus on tuning our model with more multi-modal instruction datasets and enhancing its conversation ability as a chatbot. We continue using the datasets from the second stage and add instructional datasets, including LLaVA~\citep{llava}, Flickr30k dataset~\citep{flickr30k}, our constructed mixing multi-task dataset, and the language dataset, Unnatural Instruction~\citep{honovich2022unnatural}. We give a lower data sampling ratio for the fine-grained datasets from stage-2 and a higher data sampling ratio for the new instruction datasets.

\textbf{-- LLaVA instruction data.} We add the multi-modal instruction tuning datasets, including the detailed descriptions and complex reasoning from LLaVA~\citep{llava}, with 23k and 58k data examples respectively.

\textbf{-- Flicker 30k.} After the second-stage training, our MiniGPT-v2 can effectively generate the grounded image caption. Nevertheless, these descriptions tend to be short and often cover very few number of visual objects. This is because the GRIT-20M dataset from KOSMOS-v2~\citep{peng2023kosmos} that our model was trained with, features a limited number of grounded visual objects in each caption, and our model lacks proper multi-modal instruction tuning to teach it to recognize more visual objects. To improve this, we fine-tune our model using the Flickr30k dataset~\citep{flickr30k}, which provides more contextual grounding of entities within its captions.

We prepare the Flickr30k dataset in two distinct formats for training our model to perform grounded image caption and a new task ``object parsing and grounding":

1) \textbf{Grounded image caption.} We select captions with a minimum of five grounded phrases, containing around 2.5k samples, and we directly instruct the model to produce the grounded image caption. e.g.,
\textit{a $<$p$>$wooden table$<$/p$>$\{$<$$\text{X}_{left}$$>$$<$$\text{Y}_{top}$$>$$<$$\text{X}_{right}$$>$$<$$\text{Y}_{bottom}$$>$\} in the center of the room.}

2) \textbf{Object parsing and grounding.} This new task is to parse all the objects from an input caption and then ground each object. To enable this, we use the task identifier\textit{[detection]} to differentiate this capability from other tasks. Also, we use Flickr30k to construct two types of instruction datasets: caption$\rightarrow$ grounded phrases and phrase $\rightarrow$ grounded phrase, each containing around 2.5k and 3k samples. Then we prompt our model with the instruction: \textit{[detection] description}, the model will directly parse the objects from the input image description and also ground the objects into bounding boxes.


\textbf{-- Mixing multi-task dataset.} After extensive training with single-round instruction-answer pairs, the model might not handle multiple tasks well during multi-round conversations since the context becomes more complex. To alleviate this situation, we create a new multi-round  conversation dataset by mixing the data from different tasks. We include this dataset into our third-stage model training.


\textbf{-- Unnatural instruction.} The conversation abilities of language model can be reduced after extensive vision-language training. To fix this, we add the language dataset, Unnatural Instruction~\citep{honovich2022unnatural} into our model's third-stage training for helping recover the language generation ability.

\section{Experiments}

In this section, we present experimental settings and results. We primarily conduct experiments on (detailed) image/grounded captioning, vision question answering, and visual grounding tasks, including referring expression comprehension. We present both quantitative and qualitative results. 


\begin{table*}[t!]
\small
\centering 
\resizebox{\textwidth}{!}{
\begin{tabular}{@{}lc ccccccc @{}} %
\toprule
\multirow{2}{*}{Method} & \multirow{2}{*}{Grounding} &  \multirow{2}{*}{OKVQA}  & \multirow{2}{*}{GQA} & VSR  & IconVQA & VizWiz & HM  \\  
& & & & (zero-shot) & (zero-shot)  &  (zero-shot) & (zero-shot)  \\
\midrule
Flamingo-9B & \xmark & 44.7  & - & 31.8 & -  & 28.8 & 57.0  \\
BLIP-2 (13B) & \xmark & 45.9  & 41.0 & 50.9  & 40.6 & 19.6 & 53.7  \\
InstructBLIP (13B) & \xmark & - & 49.5 & 52.1  & 44.8 & 33.4 & 57.5  \\ 
MiniGPT-4 (13B) & \xmark & 37.5   & 30.8 &  41.6  & 37.6 & - & - \\
LLaVA (13B) & \xmark & 54.4 &  41.3 & 51.2  & 43.0 & - & - \\
Shikra (13B)  & \cmark & 47.2  & - & - &  - & - & - \\
\rowcolor{Gray} 
\rowcolor{Gray}
\rowcolor{Gray}
Ours (7B)& \cmark & 56.9  & \textbf{60.3} & 60.6 & 47.7 & 32.9  & 58.2  \\
\rowcolor{Gray}


Ours (7B)-chat & \cmark & \textbf{57.8} & 60.1 & \textbf{62.9} & \textbf{51.5} & \textbf{53.6} & \textbf{58.8} \\

\bottomrule
\end{tabular}
}
\caption{\textbf{Results on multiple VQA tasks.} We report top-1 accuracy for each task. Grounding column indicates whether the model incorporates visual localization capability. The best performance for each benchmark is indicated in \textbf{bold}.}
\label{vqa_results}
\end{table*}

\begin{table*}[t!]
\small
\centering 
\resizebox{\textwidth}{!}{
\begin{tabular}{@{}lc| cccccccc c  @{}} %
\toprule
 \multirow{2}{*}{Method}&  \multirow{2}{*}{Model types} &  \multicolumn{3}{c}{RefCOCO} & \multicolumn{3}{c}{RefCOCO+} &\multicolumn{2}{c}{RefCOCOg} & \multirow{2}{*}{Avg}\\
  &&  val & test-A & test-B & val & test-A & test-B &  val & test  & \\  [0.5ex] 
\midrule
UNINEXT & \multirow{2}{*}{Specialist models} & 92.64 & 94.33 & 91.46 & 85.24 & 89.63 & 79.79 & 88.73 & 89.37  &88.90 \\
G-DINO-L & & 90.56 & 93.19 &  88.24 & 82.75 & 88.95 & 75.92 & 86.13 & 87.02 & 86.60\\
\midrule
VisionLLM-H & \multirow{6}{*}{Generalist models} & - & 86.70 & -& -& - &- &- &- & -\\
OFA-L & & 79.96 & 83.67 & 76.39 & 68.29 & 76.00 & 61.75 & 67.57 & 67.58 &  72.65\\
Shikra (7B) & &  87.01 & 90.61 & 80.24 & 81.60 & 87.36 & 72.12 & 82.27 & 82.19 & 82.93 \\ 
Shikra (13B) & &  87.83 & 91.11 & 81.81 & \textbf{82.89} & \textbf{87.79} & 74.41 & 82.64 & 83.16 & 83.96\\ 

\rowcolor{Gray}
Ours (7B) &  & \textbf{88.69} & \textbf{91.65}  & \textbf{85.33} & 79.97 & 85.12 & \textbf{74.45}& \textbf{84.44}  & \textbf{84.66} & \textbf{84.29} \\

\rowcolor{Gray}

Ours (7B)-chat & & 88.06 & 91.29 & 84.30 & 79.58 & 85.52 & 73.32 & 84.19 & 84.31 & 83.70 \\
\bottomrule
\end{tabular}
}
\caption{\textbf{Results on referring expression comprehension tasks.} Our MiniGPT-v2 outperforms many VL-generalist models including VisionLLM~\citep{wang2023visionllm}, OFA~\citep{ofa} and Shikra~\citep{shikra} and reduces the accuracy gap comparing to specialist models including UNINEXT~\citep{yan2023universal} and G-DINO~\citep{liu2023grounding}.}
\label{refcoco}
\end{table*}

\noindent \textbf{Implementation details.} Throughout the entire training process, the visual backbone of MiniGPT-v2 remains frozen. We focus on training the linear projection layer and efficient finetuning the language model using LoRA~\citep{lora}. With LoRA, we finetune $\mathcal{W}_q$ and $\mathcal{W}_v$ via low-rank adaptation. In our implementation, we set the rank, $r = 64$. We trained the model with an image resolution of 448x448 during all stages. During each stage, we use our designed multi-modal instructional templates for various vision-language tasks during the model training.

\textbf{Training and hyperparameters.} We use AdamW optimizer with a cosine learning rate scheduler to train our model.  In the initial stage, we train on 8xA100 GPUs for 400,000 steps with a global batch size of 96 and an maximum learning rate of 1e-4. This stage takes around 90 hours. During the second stage, the model is trained for 50,000 steps on 4xA100 GPUs with a maximum learning rate of 1e-5, adopting a global batch size of 64, and this training stage lasts roughly 20 hours. For the last stage, training is executed for another 35,000 steps on 4xA100 GPUs, using a global batch size of 24 and this training stage took around 7 hours, maintaining the same maximum learning rate of 1e-5.

\subsection{Quantitative Evaluation}

\noindent \textbf{Dataset and evaluation metrics.} We  evaluate our model across a range of VQA and visual grounding benchmarks. For VQA benchmarks, we consider OKVQA~\citep{schwenk2022okvqa}, GQA~\citep{hudson2019gqa}, visual spatial reasoning (VSR)~\citep{liu2023visual}, IconVQA~\citep{iconqa}, VizWiz~\citep{vizwiz}, HatefulMemes and (HM)~\citep{hateful}. For visual grounding, we evaluate our model on  RefCOCO~\citep{kazemzadeh2014referitgame} and RefCOCO+\citep{yu2016modeling}, and RefCOCOg\citep{mao2016generation} benchmarks. 


To evaluate VQA benchmarks, we use an open-ended approach with a greedy decoding strategy. We evaluate each VQA question with the following instruction template: \textit{``[vqa] {question}"}. Following the previous  method~\citep{instructblip}, we evaluate the performance by matching the model's response to the ground-truth and reporting top-1 accuracy. For visual grounding benchmarks, we use the template \textit{``[refer] give me the location of {Referring expression}"} for each referring expression comprehension question, and a predicted bounding box is considered as correct for reporting accuracy if its IOU between prediction and ground-truth is higher than 0.5.

\begin{table*}[t!]
\small
\centering 
\resizebox{0.95\textwidth}{!}{
\begin{tabular}{@{}lccccccccc @{}} %
\toprule
   &  OKVQA  & GQA & WizViz & VSR & IconVQA & HM &  Average\\  
\midrule
Ours w/o task identifier  & 50.5  & 53.4 &  28.6 & 57.5 & 44.8 & 56.8 & 48.6 \\
Ours  & \textbf{52.1} & \textbf{54.6} & \textbf{29.4}& \textbf{59.9} &  \textbf{45.6} & \textbf{57.4} & \textbf{49.8}  \\
\bottomrule
\end{tabular}
}
\caption{Task identifier ablation study on VQA benchmarks. With task identifier during the model training can overall improve VQA performances from multiple VQA benchmarks}
\label{ablation}
\end{table*}

\begin{table*}[t!]
\centering 
\resizebox{0.6\textwidth}{!}{
\begin{tabular}{@{}lc cc @{}} %
\toprule
Method & $\text{CHAIR}_I$ $\downarrow$ &  $\text{CHAIR}_S$ $\downarrow$ & Len  \\  
\midrule
MiniGPT-4 & 9.2& 31.5 & 116.2 \\
mPLUG-Owl  & 30.2 & 76.8  & 98.5 \\
LLaVA & 18.8 &  62.7&  90.7 \\
MultiModal-GPT & 18.2 & 36.2 & 45.7 \\
\rowcolor{Gray}
MiniGPT-v2 (long) & 8.7 & 25.3 & 56.5 \\
\rowcolor{Gray}
MiniGPT-v2 (grounded) & 7.6  & 12.5 & 18.9 \\
\rowcolor{Gray}
MiniGPT-v2 (short) & \textbf{4.4} & \textbf{7.1} & \textbf{10.3}\\
\bottomrule
\end{tabular}
}
\caption{\textbf{Results on hallucination.} We evaluate the hallucination of MiniGPT-v2 with different instructional templates and output three versions of captions for evaluation. For the ``long" version, we use the prompt \textit{generate a brief description of the given image}. For the ``grounded" version, the instruction is \textit{[grounding] describe this image in as detailed as possible}. For the ``short" version, the prompt is \textit{[caption] briefly describe the image}. 
}
\label{halluciniation}
\end{table*}

\noindent \textbf{Visual question answering results.} Table~\ref{vqa_results} presents our experimental results on multiple VQA benchmarks. Our results compare favorably to baselines including MiniGPT-4~\citep{zhu2023minigpt}, Shikra~\citep{shikra}, LLaVA~\citep{llava}, and InstructBLIP~\citep{instructblip} across all the VQA tasks. For example, on QKVQA, our MiniGPT-v2 outperforms MiniGPT-4, Shikra, LLaVA, and BLIP-2 by 20.3\%, 10.6\%, 3.4\%, and 11.9\%. These results indicate the strong visual question answering capabilities of our model. Furthermore, we find that our MiniGPT-v2 (chat) variant shows higher performance than the version trained after the second stage. On OKVQA,  VSR, IconVQA, VizWiz, and HM, MiniGPT-v2 (chat) outperforms MiniGPT-v2 by 0.9\%,  2.3\%, 4.2\%, 20.7\%, and 0.6\%. We believe that the better performance can be attributed to the improved language skills during the third-stage training, which is able to benefit visual question comprehension and response, especially on VizWiz with 20.7\% top-1 accuracy increase.

\noindent \textbf{Referring expression comprehension results.} Table \ref{refcoco} compares our model to baselines on REC benchmarks. Our MiniGPT-v2 shows strong REC performance on RefCOCO, RefCOCO+, and RefCOCOg, performing better than  other vision-language generalist models. MiniGPT-v2 outperforms OFA-L~\citep{ofa} by over 8\% accuracy across all tasks of RefCOCO/RefCOCO+/RefCOCOg. Compared with a strong baseline, Shikra (13B) ~\citep{shikra}, our model still shows better results, e.g., 84.29\% vs 83.96\% accuracy in average. These results provide direct evidence for  the competing visual grounding capabilities of MiniGPT-v2. Although our model underperforms specialist models, the promising performance indicates its growing competence in visual grounding.

\textbf{Ablation on task identifier.} We conduct ablation studies on the effect of the task identifier on the performance of MiniGPT-v2. We compare our model with the variant without using task identifiers on VQA benchmarks. Both models were trained on 4xA100 GPUs for 24 hours with an equal number of training steps for multiple vision-language tasks. Results in Table \ref{ablation} demonstrate the performance on multiple VQA benchmarks and consistently show that token identifier training benefits the overall performance of MiniGPT-v2. Specifically, our MiniGPT-v2 with task-oriented instruction training achieves 1.2\% top-1  accuracy improvement on average. 
These ablation results can validate the clear advantage of adding task identifier tokens and support the use of multi-task identifiers for multi-task learning efficiency.

\begin{figure}[t!]
    \centering
    \includegraphics[width=\textwidth]{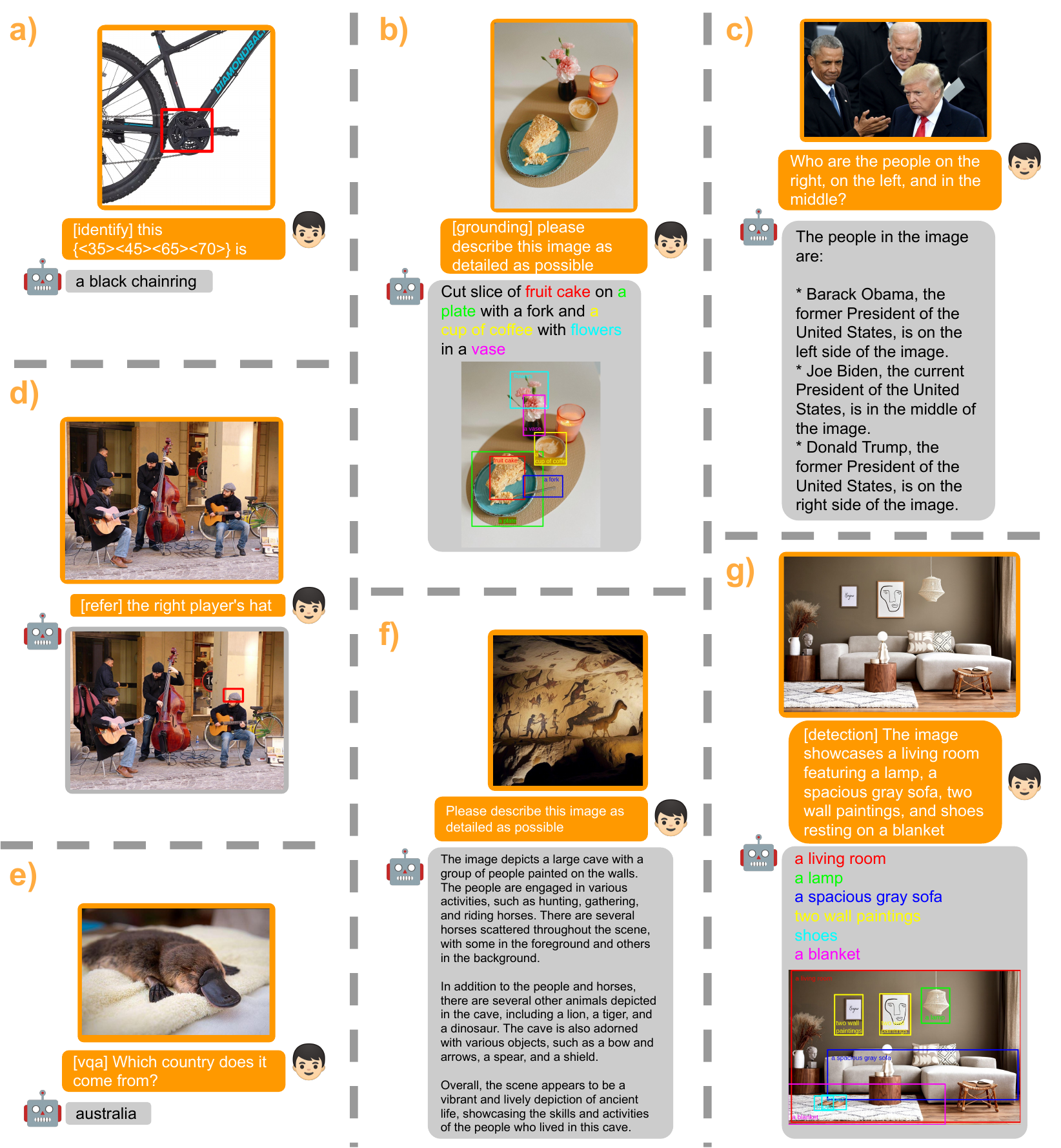}
    \caption{\textbf{Examples for various multi-modal capabilities of MiniGPT-v2.} We showcase that our model is capable of completing multiple tasks such as referring expression comprehension, referring expression generation, detailed grounded image caption, visual question answering, detailed image description, and directly parsing phrase and grounding from a given input text. }
    \label{fig:demo}
\end{figure}





\noindent \textbf{Hallucination.} We measure the hallucination of our model on image description generation and compare the results with other vision-language baselines, including MiniGPT-4~\citep{zhu2023minigpt}, mPLUG-Owl~\citep{mplugowl}, LLaVA~\citep{llava}, and MultiModal-GPT~\citep{gong2023multimodal}. Following the methodology from \citep{li2023evaluating}, we use CHAIR~\citep{rohrbach2018object} to assess hallucination at both object and sentence levels. As shown in Table \ref{halluciniation}, we find that our MiniGPT-v2 tends to generate the image description with reduced hallucination compared to other baselines. We have evaluated three types of prompts in MiniGPT-v2. 
First, we use the prompt \textit{generate a brief description of the given image} without any specific task identifier which tends to produce more detailed image descriptions. Then we provide the instruction prompt \textit{[grounding] describe this image in as detailed as possible} for evaluating grounded image captions. Lastly, we prompt our model with \textit{[caption] briefly describe the image}. With these task identifiers, MiniGPT-v2 is able to produce a variety of image descriptions with different levels of hallucination. As a result, all these three instruction variants have lower hallucination than our baseline, especially with the task specifiers of \textit{[caption]} and \textit{[grounding]}.


\subsection{Qualitative Results}
We now provide the qualitative results for a complementary understanding of our model's multi-modal capabilities. Some examples can be seen in Fig. \ref{fig:demo}. Specifically, we demonstrated various abilities in the examples including a) object identification; b) detailed grounded image captioning; c) visual question answering; d) referring expression comprehension; e) visual question answering under task identifier; f) detailed image description; g) object parsing and grounding from an input text. More qualitative results can be found in the Appendix. These results demonstrate that our model has competing vision-language understanding capabilities. Moreover, notice that we train our model only with a few thousand of instruction samples on object parsing and grounding tasks at the third-stage, and our model can effectively follow the instructions and generalize on the new task. This indicates that our model has the flexibility to adapt on many new tasks.

Note that our model still occasionally shows hallucinations when generating the image description or visual grounding. e.g., our model may sometimes produce descriptions of non-existent visual objects or generate inaccurate visual locations of grounded objects. We believe training with more high-quality image-text aligned data and integrating with a stronger vision backbone or large language model hold the potential for alleviating this issue.

\section{Conclusion}

In this paper, we introduce MiniGPT-v2, a multi-modal LLM that can serve as a unified interface for various vision-language multi-tasking learning. To develop a single model capable of handling multiple vision-language tasks, we propose using distinct identifiers for each task during the training and inference. These identifiers help our model easily differentiate various tasks and also improve learning efficiency. Our MiniGPT-v2  achieves state-of-the-art results across many visual question answering and referring expression comprehension benchmarks. We also found that our model can efficiently adapt to new vision-language tasks, which suggests that MiniGPT-v2 has many potential applications in the vision-language community. 



\clearpage

{\small
\bibliographystyle{ieee_fullname}
\bibliography{egbib}
}

\newpage
\appendix
\section{Appendix}

In the supplementary, we provide more qualitative results that are generated from our model to demonstrate the vision-language multi-tasking capabilities.

\subsection{Instruction template for various vision-language tasks}

\textbf{RefCOCO/RefCOCO+/RefCOCOg:} \textit{[refer] give me the location of {question}}

\textbf{VizWiz:} \textit{[vqa] Based on the image, respond to this question with a single word or phrase: {question}, and reply 'unanswerable' when the provided information is insufficient}

\textbf{Hateful Meme:} \textit{[vqa] This is an image with: {question} written on it. Is it hateful? Answer:}

\textbf{VSR:} \textit{[vqa] Based on the image, is this statement true or false? {question}}

\textbf{IconQA, GQA, OKVQA:} \textit{[vqa] Based on the image, respond to this question with a single word or phrase: {question}}

\subsection{Additional Qualitative Results}

\begin{figure}[h]
    \centering
    \includegraphics[width=\textwidth]{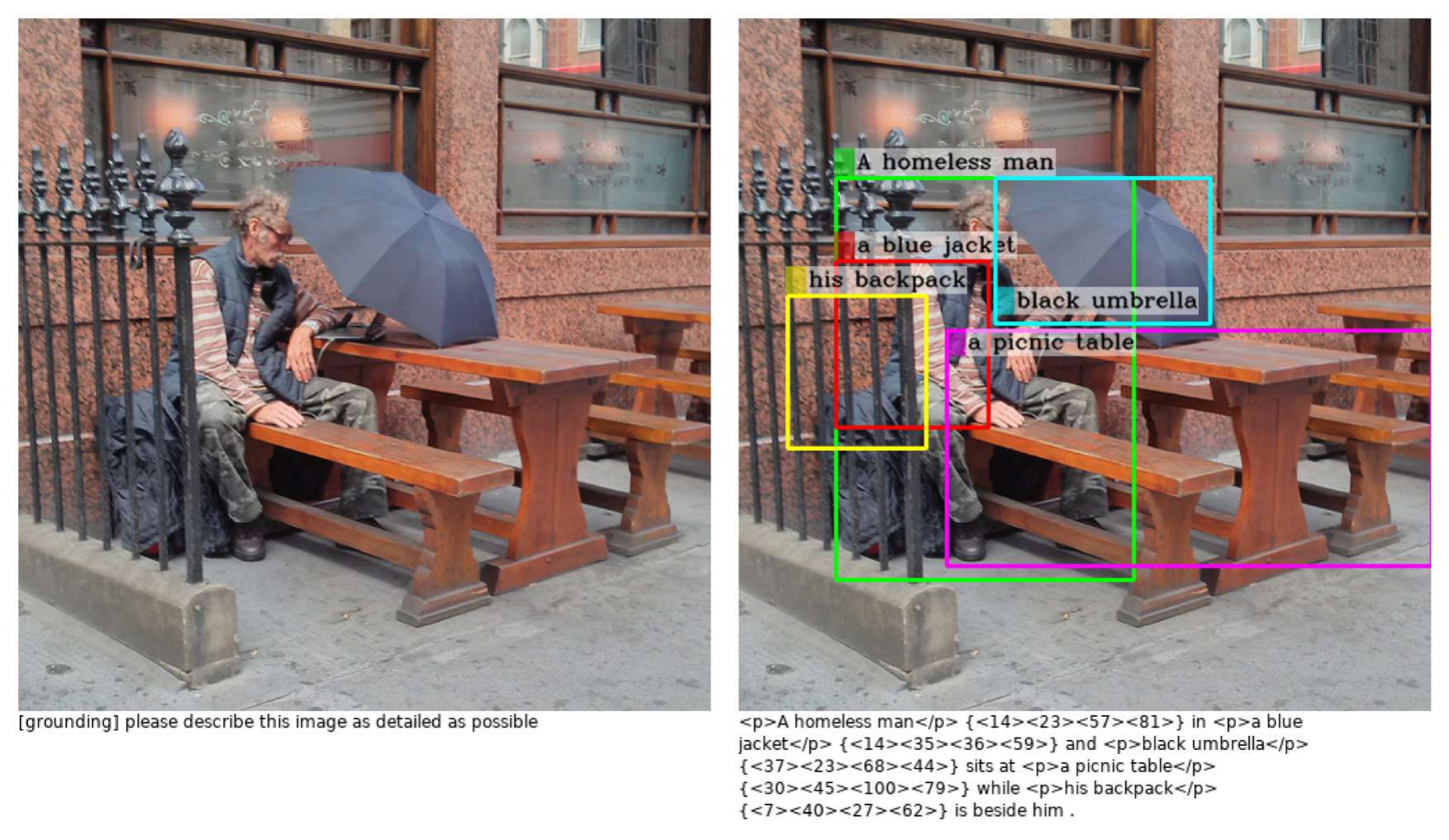}
    \caption{Detail grounded image caption example.}
    \label{g1}
\end{figure}

\begin{figure}[h]
    \centering
    \includegraphics[width=\textwidth]{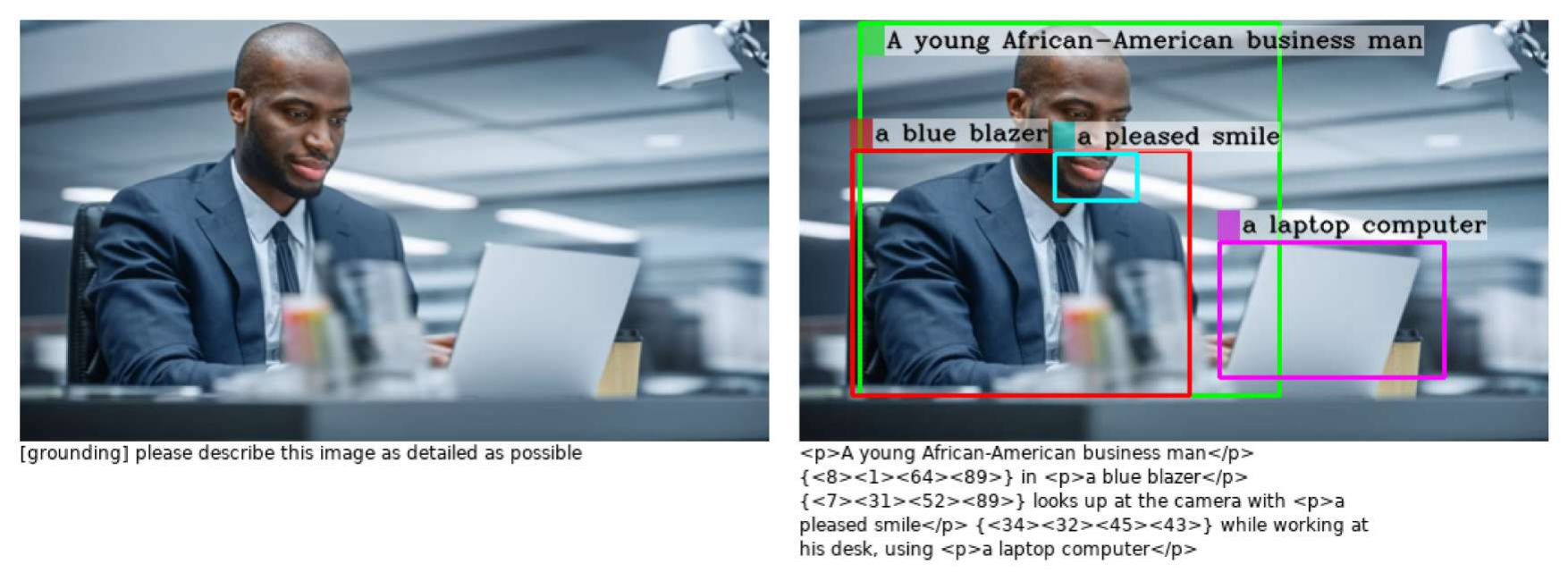}
        \caption{Detail grounded image caption example}
        \label{g2}
\end{figure}

\begin{figure}[h]
    \centering
    \includegraphics[width=\textwidth]{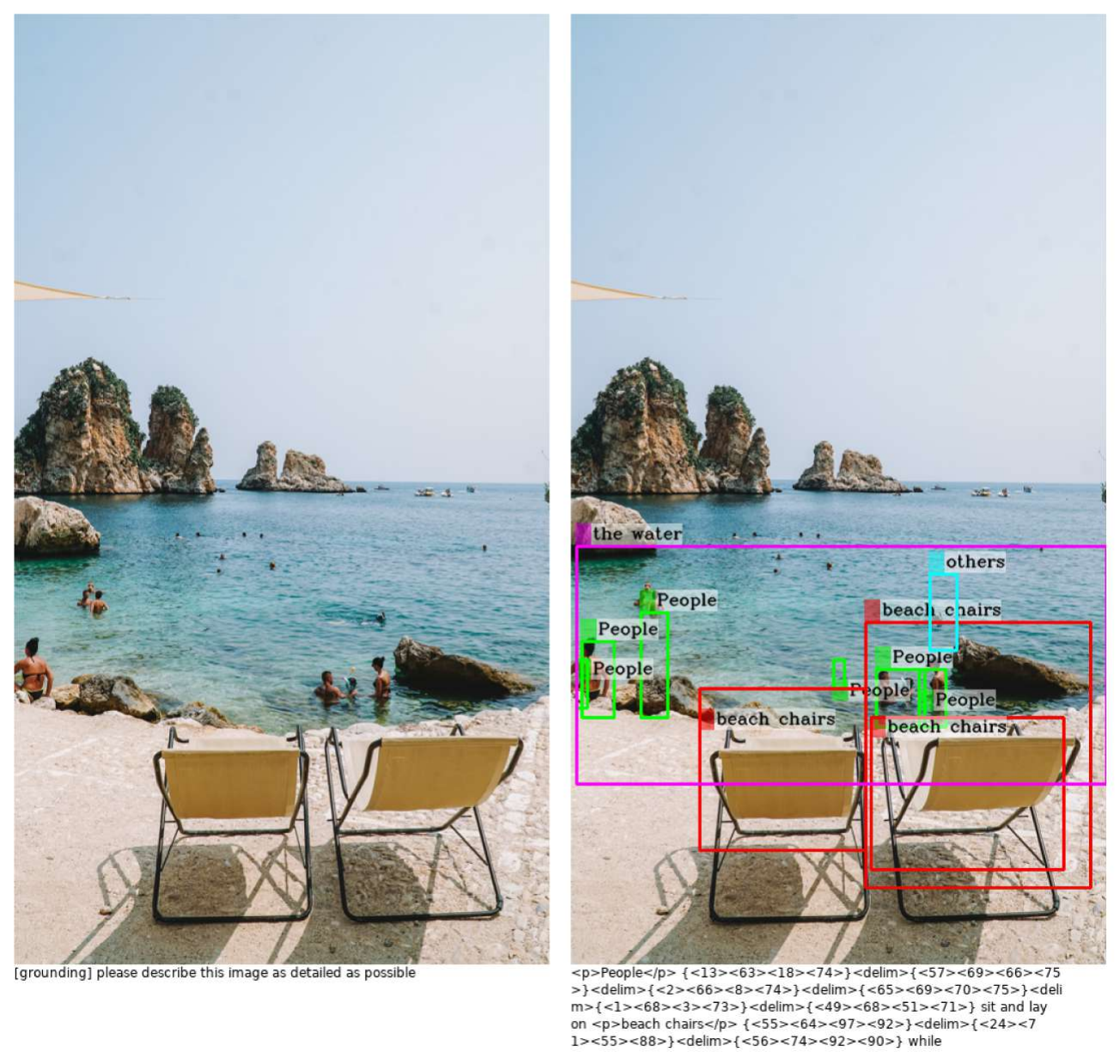}
        \caption{Detail grounded image caption example}
    \label{g3}
\end{figure}

\begin{figure}[h]
    \centering
    \includegraphics[width=\textwidth]{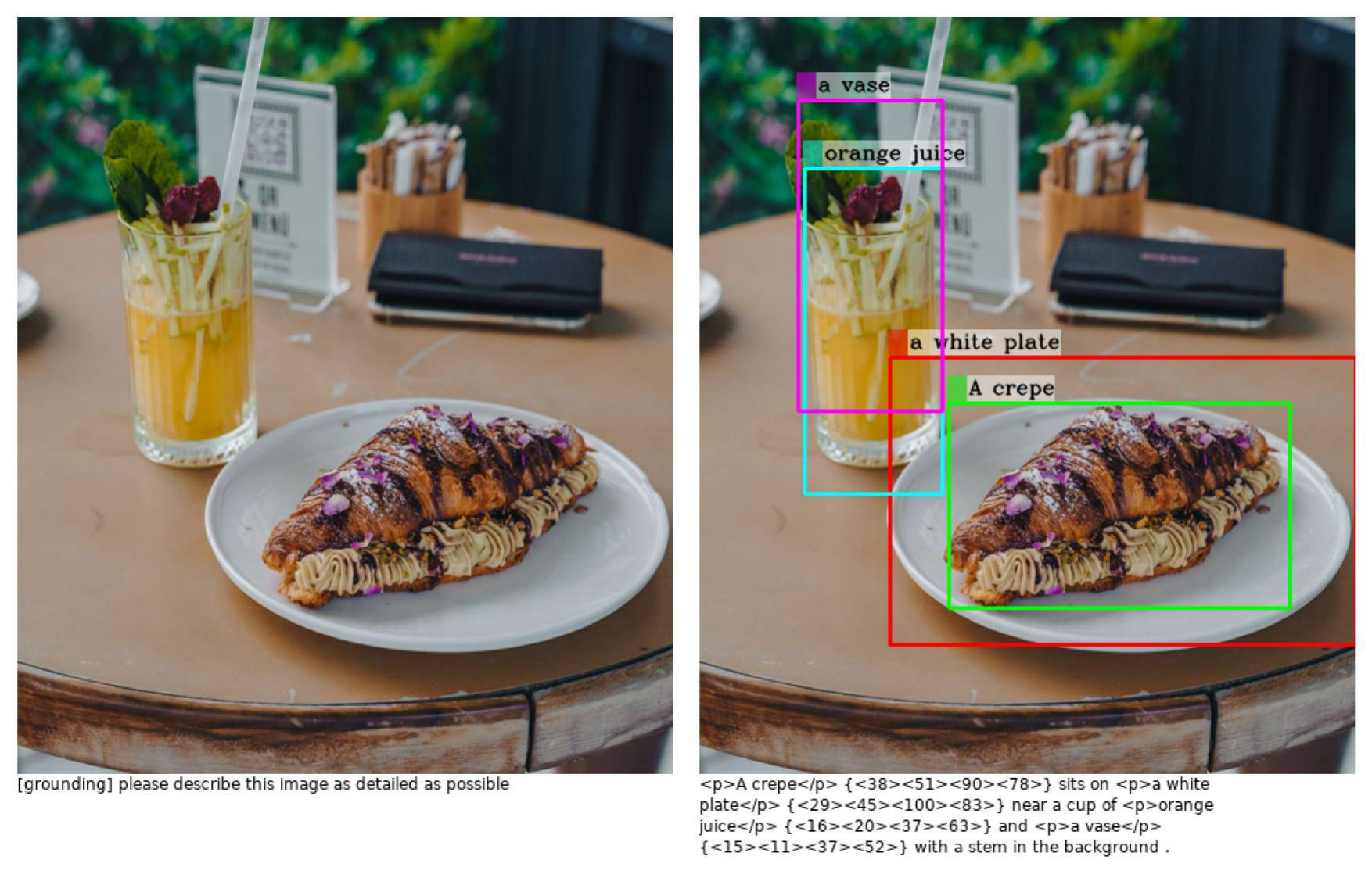}
    \caption{Detail grounded image caption example}
    \label{g4}
\end{figure}

\begin{figure}[h]
    \centering
    \includegraphics[width=\textwidth]{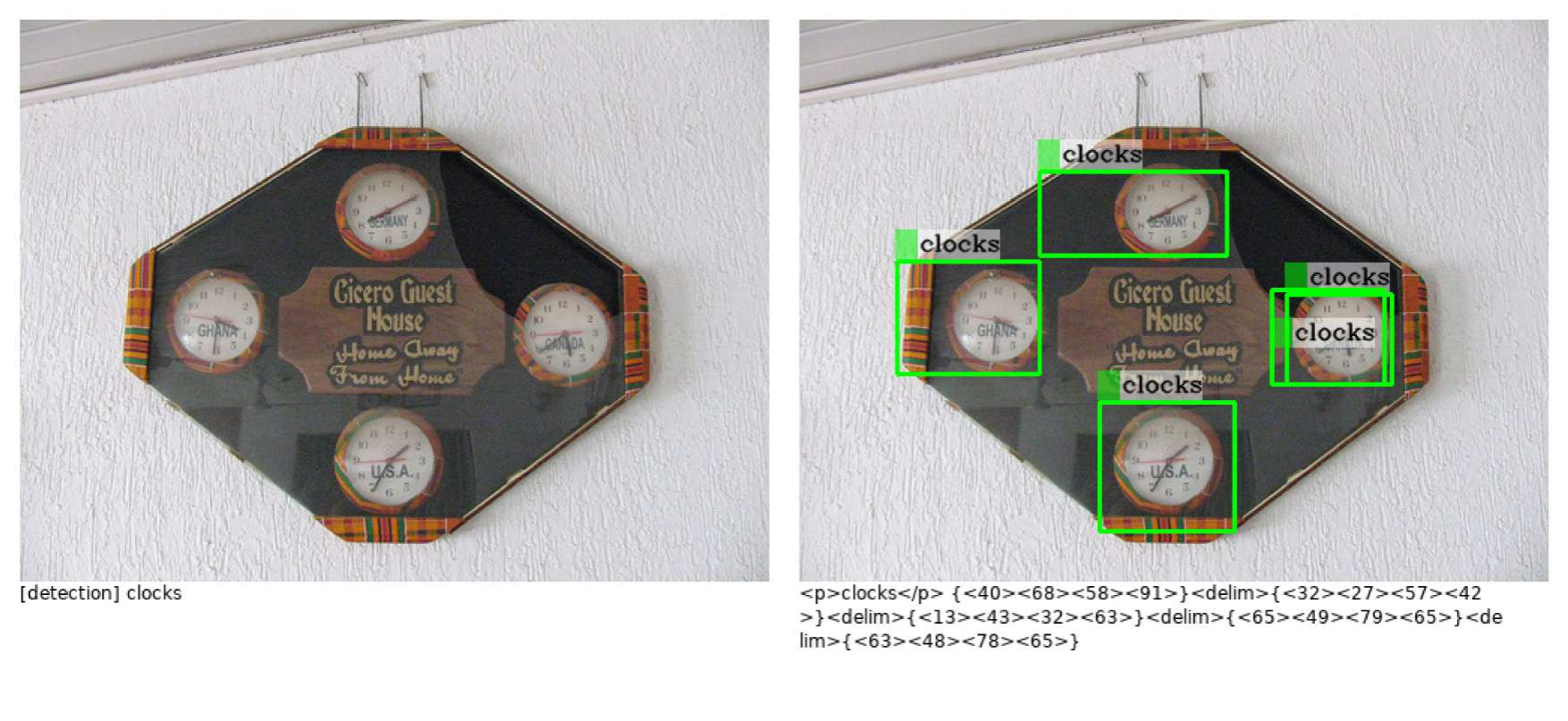}  
    \caption{Object parsing and grounding example}
    \label{d1}
\end{figure}

\begin{figure}[h]
    \centering
    \includegraphics[width=\textwidth]{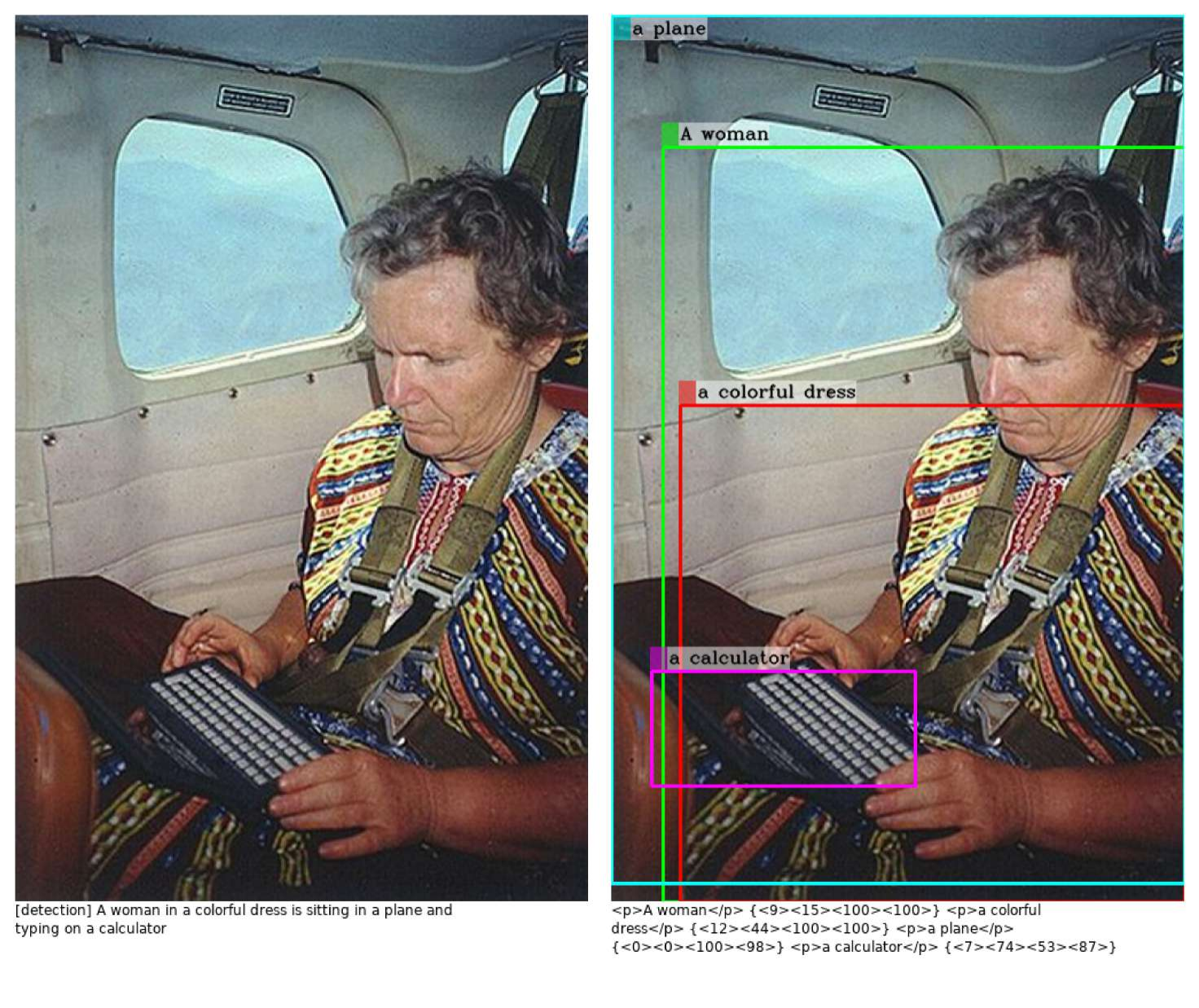}
    \caption{Object parsing and grounding example}
    \label{d2}
\end{figure}

\begin{figure}[h]
    \centering
    \includegraphics[width=\textwidth]{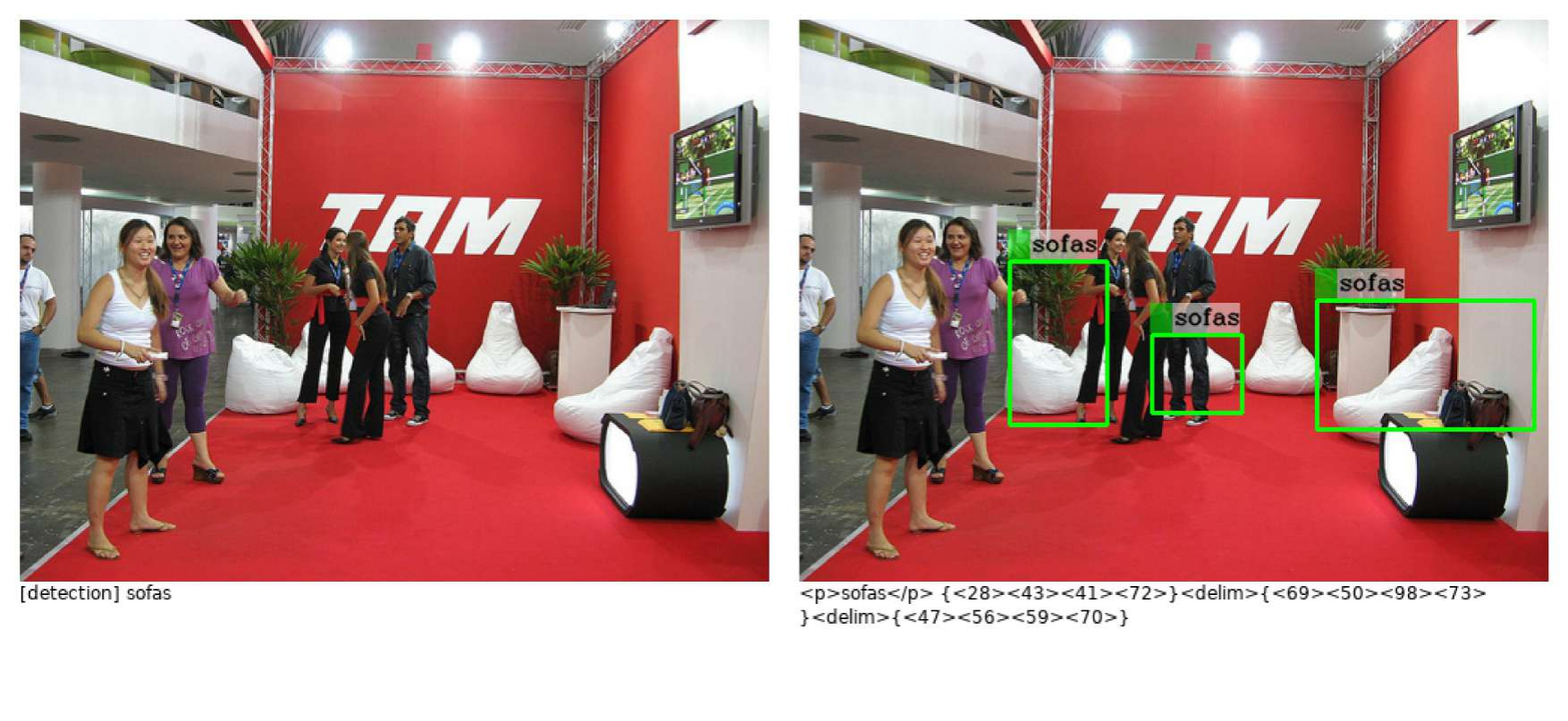}
    \caption{Object parsing and grounding example}
    \label{d3}
\end{figure}

\begin{figure}[h]
    \centering
    \includegraphics[width=\textwidth]{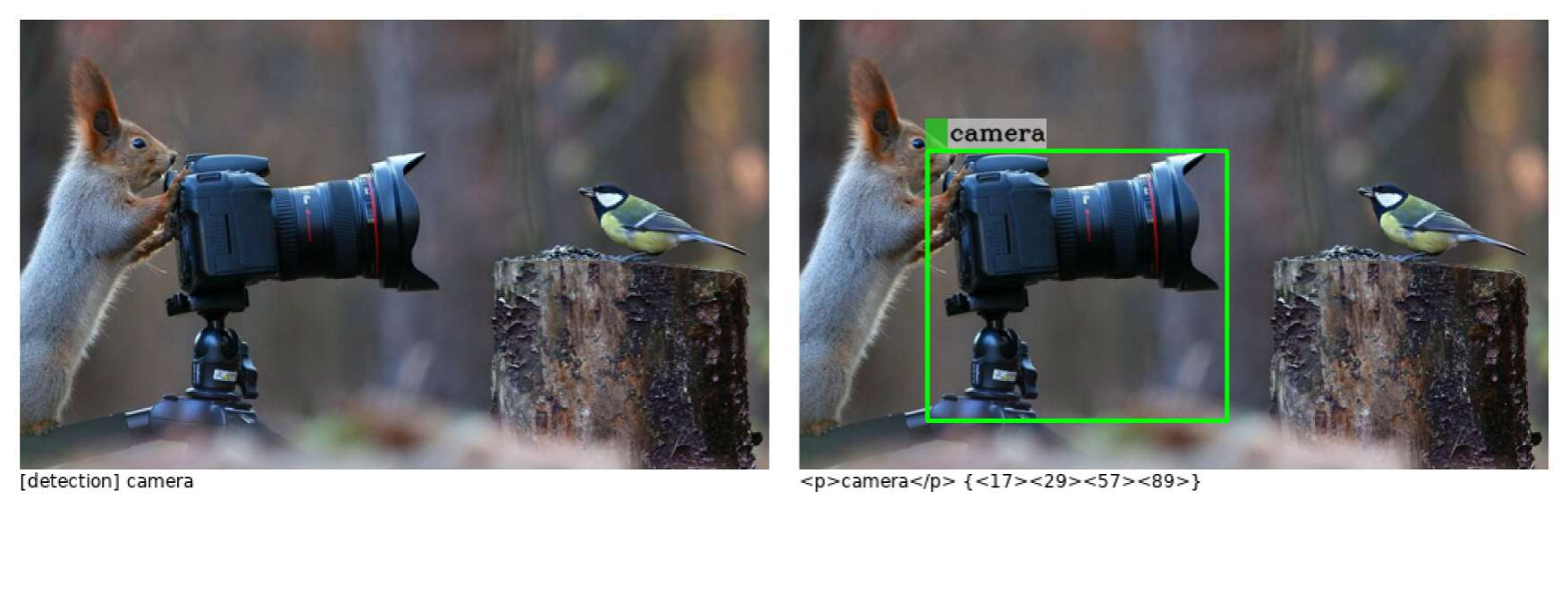}
    \caption{Object parsing and grounding example}
    \label{d4}
\end{figure}

\begin{figure}[h]
    \centering
    \includegraphics[width=\textwidth]{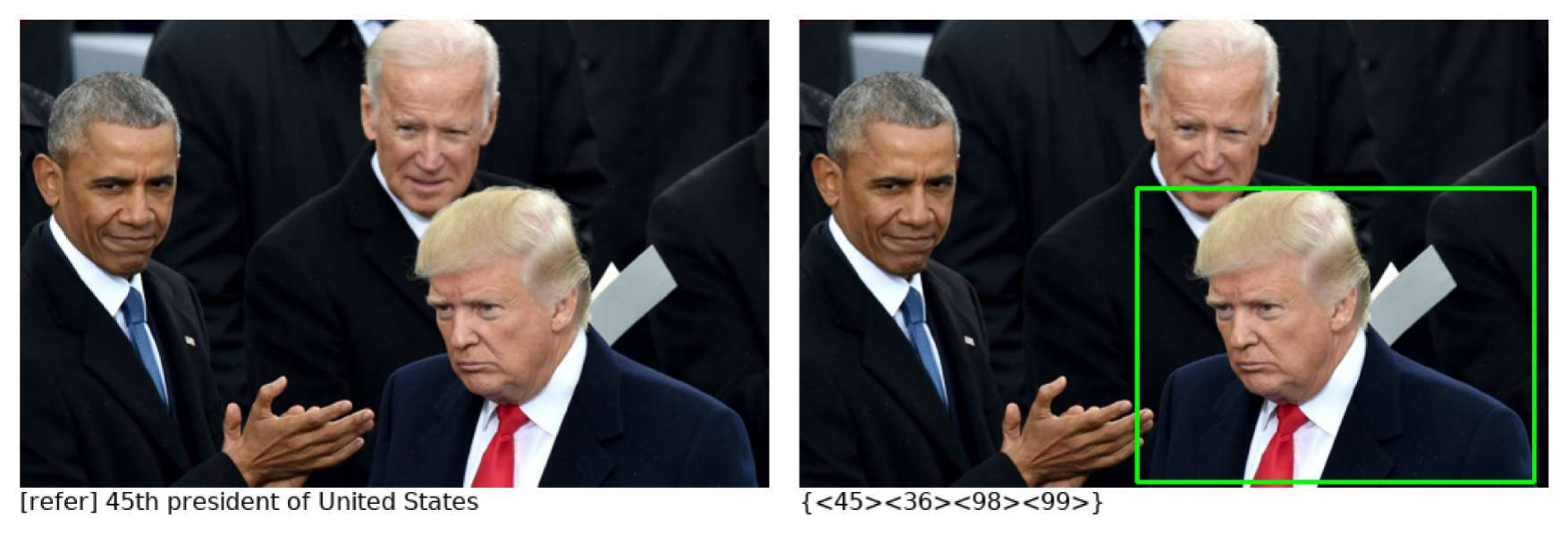}
    \caption{Referring expression comprehension example}
    \label{r1}
\end{figure}

\begin{figure}[h]
    \centering
    \includegraphics[width=\textwidth]{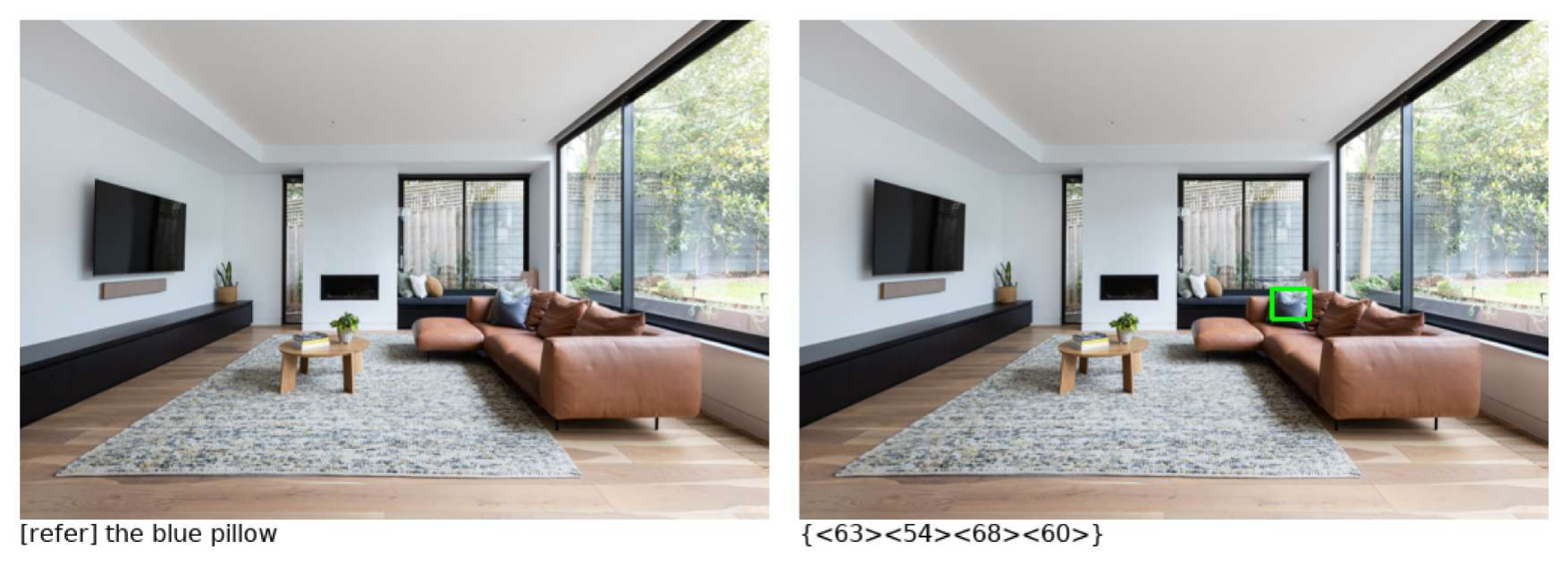}
    \caption{Referring expression comprehension example}
    \label{r2}
\end{figure}

\begin{figure}[h]
    \centering
    \includegraphics[width=\textwidth]{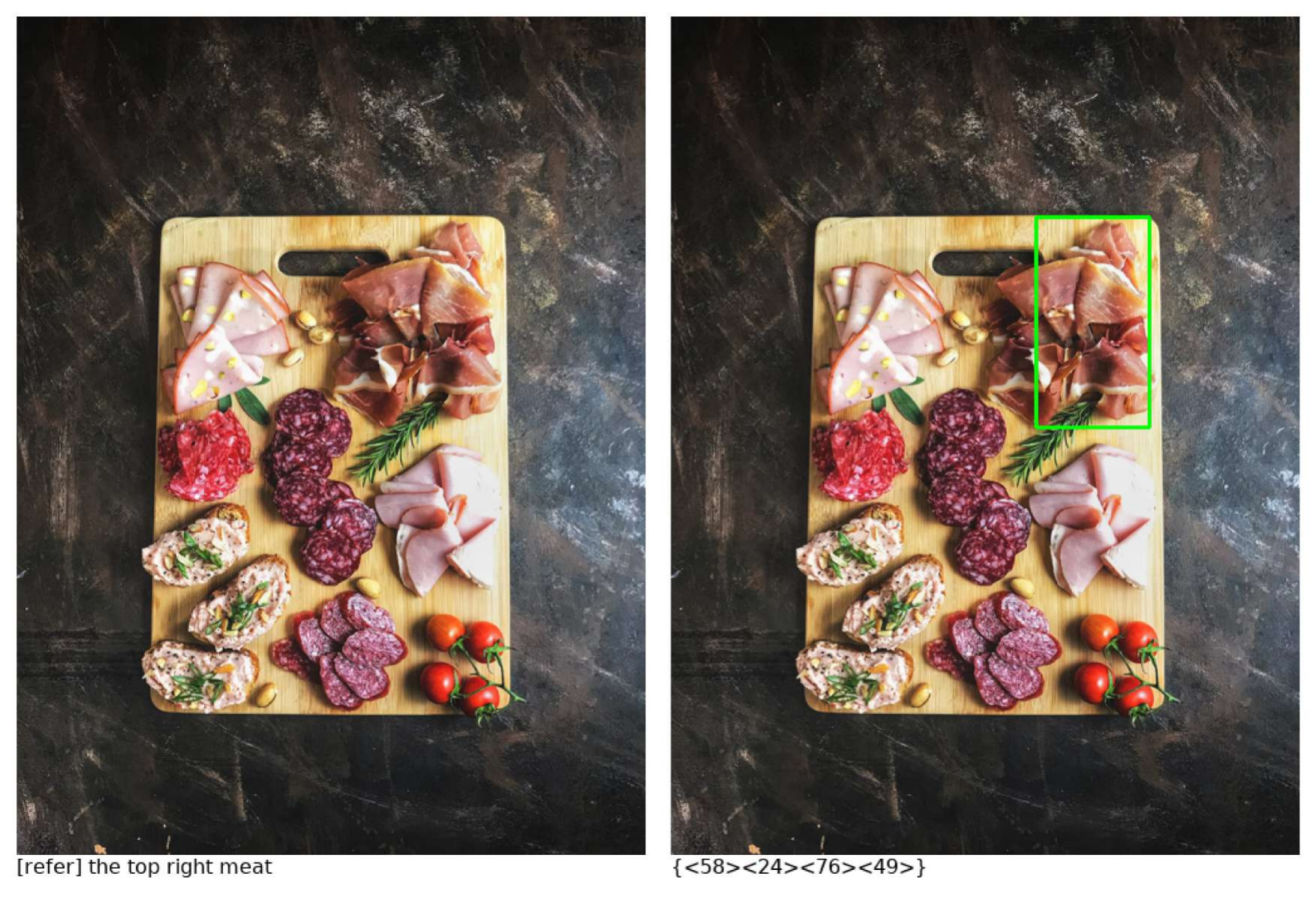}
    \caption{Referring expression comprehension example}
    \label{r3}
\end{figure}

\begin{figure}[h]
    \centering
    \includegraphics[width=\textwidth]{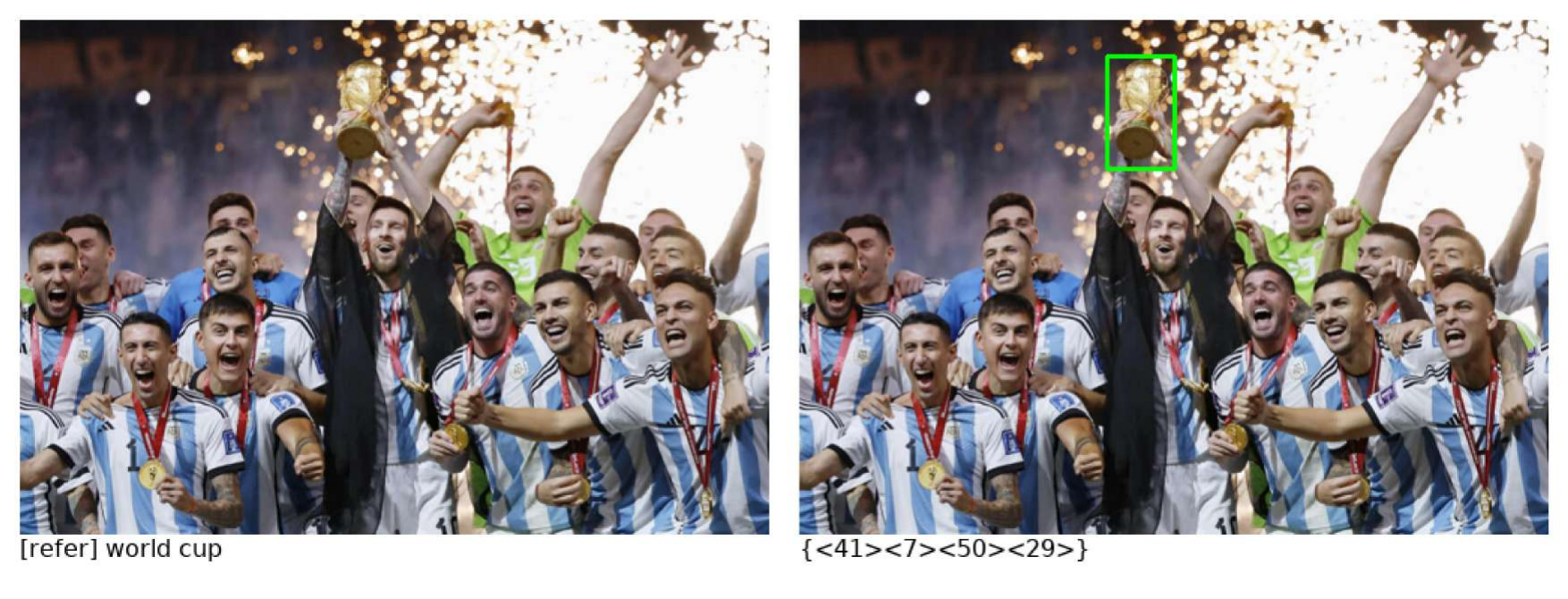}
    \caption{Referring expression comprehension example}
    \label{r4}
\end{figure}

\begin{figure}[h]
    \centering
    \includegraphics[width=\textwidth]{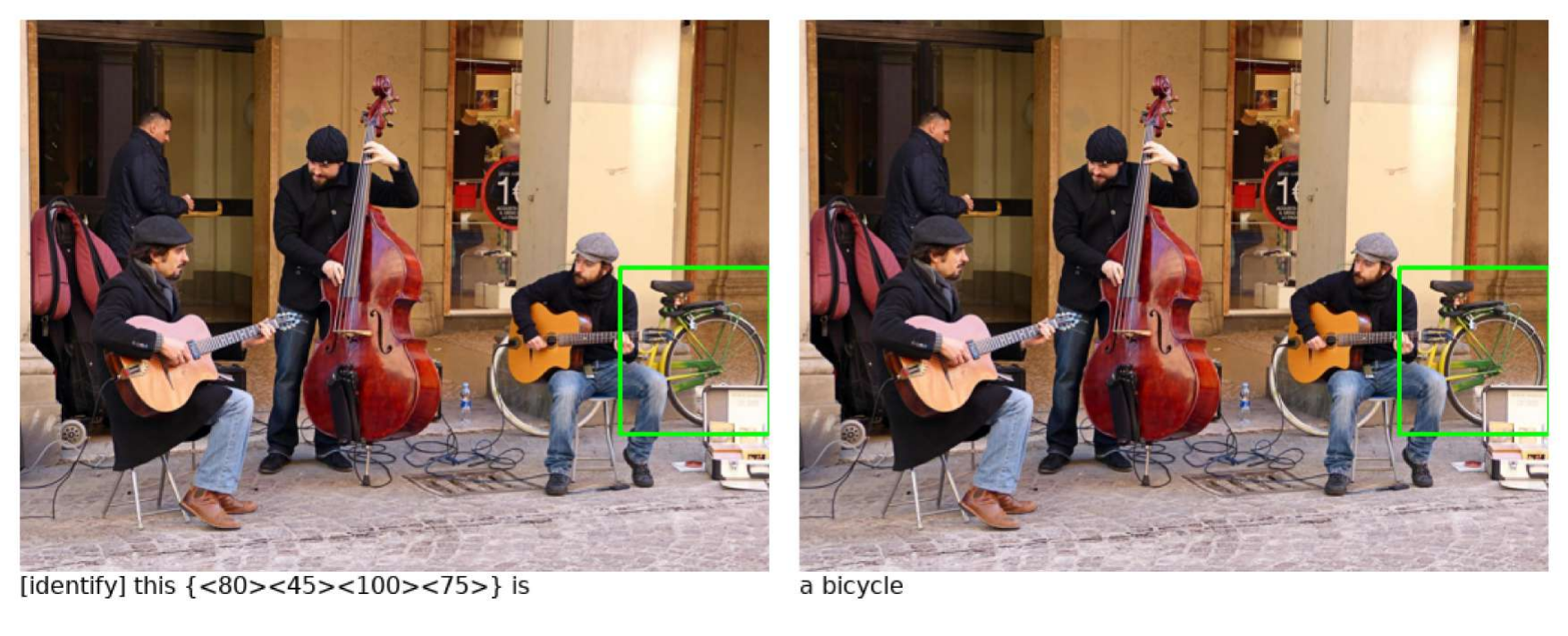}
    \caption{object identification example}
    \label{i1}
\end{figure}

\begin{figure}[h]
    \centering
    \includegraphics[width=\textwidth]{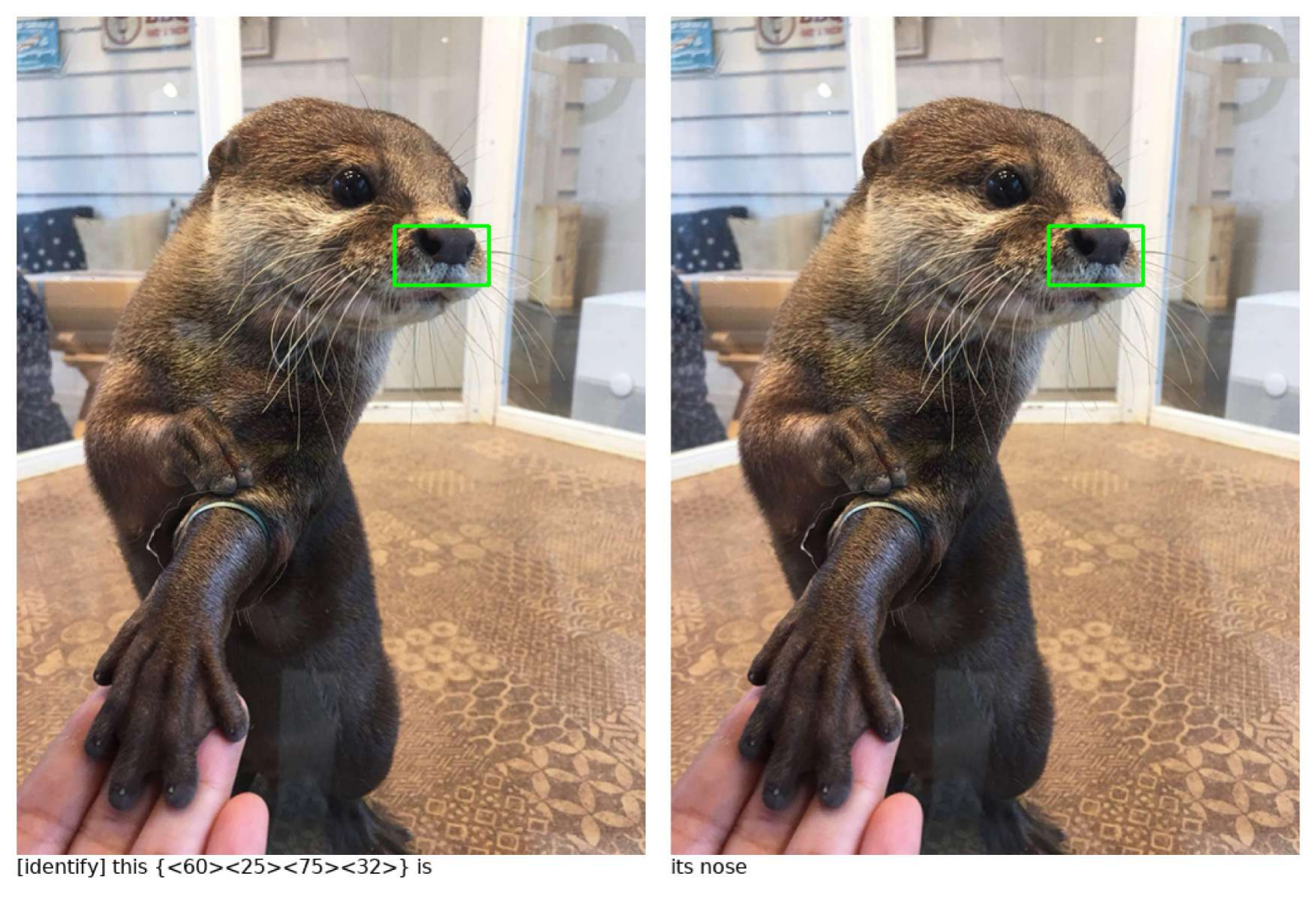}
    \caption{object identification example}
    \label{i2}
\end{figure}

\begin{figure}[h]
    \centering
    \includegraphics[width=\textwidth]{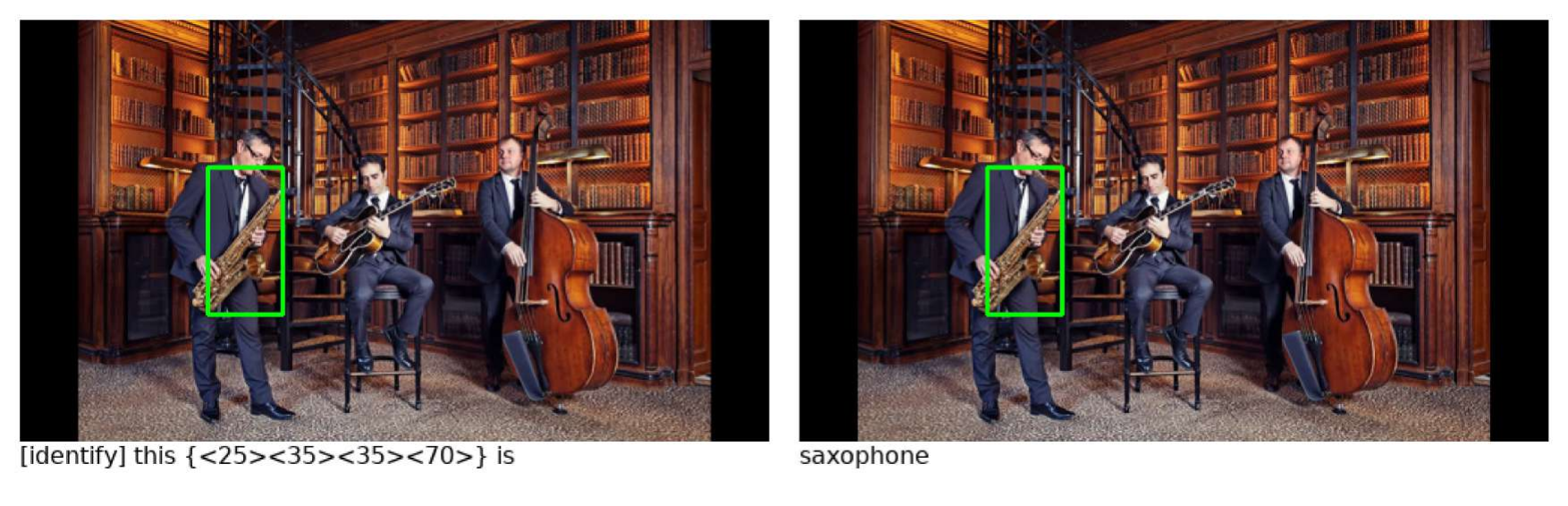}
    \caption{object identification example}
    \label{i3}
\end{figure}

\begin{figure}[h]
    \centering
    \includegraphics[width=\textwidth]{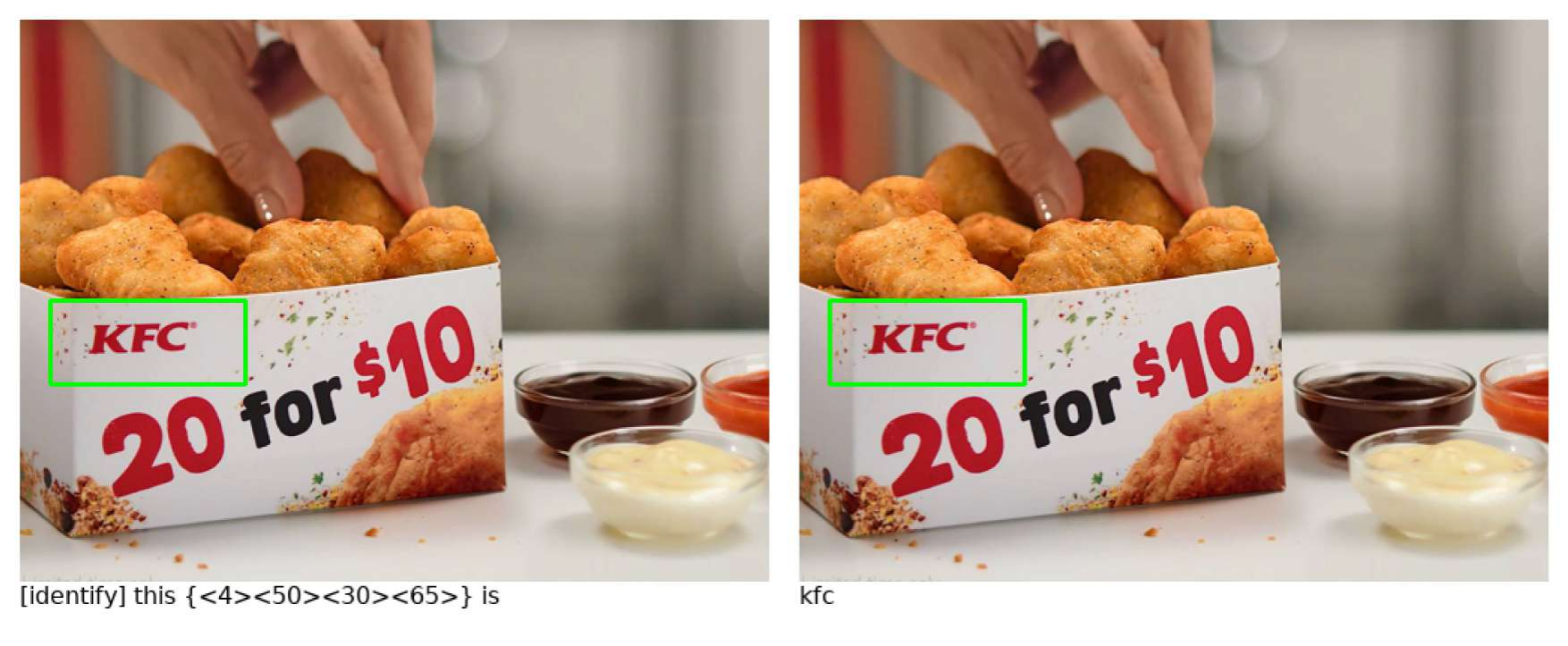}
    \caption{object identification example}
    \label{i4}
\end{figure}

To study how well our model is able to take visual input and answer questions based on task-oriented identifier, we use our model to perform multiple vision-language tasks including  grounded image captioning in Fig. \ref{g1}, Fig. \ref{g2}, Fig. \ref{g3} and Fig. \ref{g4}; Object parsing and grounding in Fig. \ref{d1}, Fig. \ref{d2}, Fig. \ref{d3} and Fig. \ref{d4}; Referring expression comprehension in Fig. \ref{r1}, Fig. \ref{r2}, Fig. \ref{r3} and Fig. \ref{r4}; Object identification in Fig. \ref{i1}, Fig. \ref{i2}, Fig. \ref{i3} and Fig. \ref{i4}. 



For each task, we share 4 examples for showing the vision-language capabilities of our model. The results in the demo provide direct evidence for the competing visual understanding capabilities of MiniGPT-v2 on multiple vision-language tasks. For example, in the cases of grounded caption, our model is able to give correct grounded image caption with detailed spatial locations of objects. In the cases of identify, the model also generates our expected object names. MiniGPT-v2 can understand the new scenes and follow the question identifier to respond. But we also need to note that our model still has some hallucination e.g., In Fig. \ref{g3}, several persons are not grounded accurately, and in Fig. \ref{g4}, there does not exist a vase in the image.


\end{document}